\title{\LARGE \bf
Learning Setup Policies: Reliable Transition
\\ Between Locomotion Behaviours
}

\author{Brendan Tidd$^{1,2}$, Nicolas Hudson$^{2}$, Akansel Cosgun$^{3}$ and J\"{u}rgen Leitner$^{4}$
\thanks{$^{1}$Queensland University of Technology (QUT), Australia}%
\thanks{$^{2}$Robotics and Autonomous Systems Group, CSIRO, Pullenvale, QLD 4069, Australia \newline Email: brendan.tidd@data61.csiro.au}
\thanks{$^{3}$Monash University, Australia}%
\thanks{$^{4}$LYRO Robotics Pty Ltd, Australia}%
}

\begin{document}

\maketitle
\thispagestyle{empty}
\pagestyle{empty}

\begin{abstract}
\label{sec:abstract}
Dynamic platforms that operate over many unique terrain conditions typically require many behaviours. To transition safely, there must be an overlap of states between adjacent controllers. We develop a novel method for training setup policies that bridge the trajectories between pre-trained Deep Reinforcement Learning (DRL) policies. We demonstrate our method with a simulated biped traversing a difficult jump terrain, where a single policy fails to learn the task, and switching between pre-trained policies without setup policies also fails. We perform an ablation of key components of our system, and show that our method outperforms others that learn transition policies. We demonstrate our method with several difficult and diverse terrain types, and show that we can use setup policies as part of a modular control suite to successfully traverse a sequence of complex terrains. We show that using setup policies improves the success rate for traversing a single difficult jump terrain (from 1.5$\%$ success rate without setup policies to 82$\%$), and traversing a random sequence of difficult obstacles (from 1.9$\%$ without setup policies to 71.2$\%$).

\end{abstract}

\section{Introduction}
\label{sec:introduction}

Bipedal robots have the capability to cover a span of terrains that humans can traverse. This makes bipeds an appealing locomotion option for robots in human environments. It is impractical, however, to design a single controller that works over all required environments, particularly if all conditions are not known a priori. For robots to be able to perform a number of diverse behaviours, it is desirable for a control suite to be modular, where any number of behaviours can be added with minimal retraining.

 
Deep Reinforcement Learning (DRL) is an appealing method for developing visuo-motor behaviours for legged platforms. However, such DRL policies are usually trained on the terrain they are expected to operate, though it is unlikely that the agent would have access to a complete set of terrain conditions during policy training before deployment. Controllers for dynamic platforms are typically developed and tuned with safety harnesses and human supervisors in order to minimise potential damage to the robot and the environment. These restrictions limit the development of controllers over an exhaustive collection of terrain conditions. Furthermore, modifying controllers that have already been tuned is costly, for example DRL methods typically require retraining if new terrains are introduced. We propose training a \textit{setup policy} to transition between two pre-trained policies targeting different terrains.

\begin{figure}[tb!]
\centering
\includegraphics[width=\columnwidth]{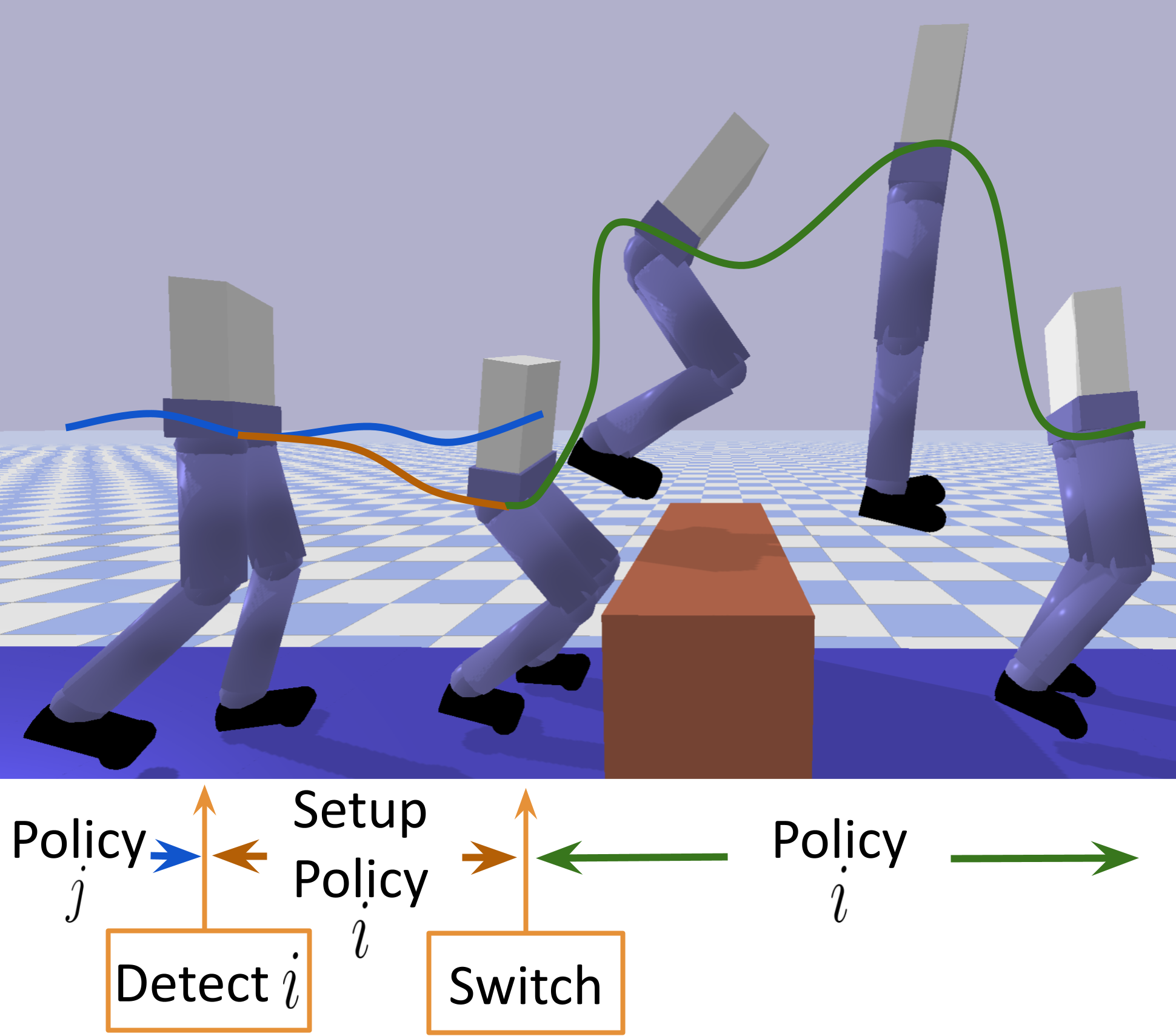}
\hfill
\caption{As the robot approaches a difficult terrain artifact, it must move into position for target policy $i$. The trajectory of the walking policy $j$ (shown with a blue line), does not intersect with the trajectory of the target policy $i$ (green line). Our setup policy provides a trajectory that prepares the robot for the upcoming target policy (orange line).}
\label{fig:fig1}
\end{figure}

A modular control suite for a dynamic platform requires careful design. Simply switching between controllers may result in the robot falling over if controllers are switched when the robot is not in a safe region for the subsequent controller. Figure~\ref{fig:fig1} shows our primary contribution: a setup policy that prepares the robot for the necessary controller to traverse the upcoming terrain obstacle. With our method, we can transition between existing behaviours, allowing pre-trained DRL policies to be combined. 


Our contributions are as follows:
\begin{itemize}
    \item Using pre-trained policies, we develop \textbf{setup policies} that significantly improve the success rate from transitioning from a default walking policy to a target policy (from 1.5$\%$ success rate without setup policies to 82$\%$ for a difficult 0.5m jump). The setup policy also learns when to switch to the target policy. 
    \item We introduce a novel reward, called \textbf{Advantage Weighted Target Value}, guiding the robot towards the target policy.
    \item We show that we can use setup policies with several difficult terrain types, allowing for a modular control suite, combining behaviours that have been trained separately without any retraining of low level policies.
\end{itemize}


\section{Related Work}
\label{sec:related_work}

Deep reinforcement learning (DRL) has demonstrated impressive results for locomotion tasks in recent works~\cite{heess_emergence_2017, peng_deeploco_2017, peng_deepmimic_2018, xie_allsteps_2020}. Typically DRL policies optimise a single reward function, as new environments and behaviours are introduced costly retraining is required. Single policies have demonstrated locomotion over various terrains for simple 2D cases \cite{song_recurrent_2018}, and impressive quadruped behaviours \cite{lee_learning_2020, rudin_learning_2021}, however, for many scenarios multiple policies are often required. Furthermore, developing a single policy to perform multiple behaviours can degrade the performance of each behaviour~\cite{lee_robust_2019}.

Hierarchical reinforcement learning (HRL) offers flexibility through architecture, by training all segments of the hierarchy concurrently~\cite{sutton_between_1999, bacon_option-critic_2018, frans_meta_2018,peng_deeploco_2017}, or training parts of the hierarchy separately~\cite{merel_hierarchical_2019},~\cite{lee_composing_2019}.  When trained together, HRL can improve task level outcomes, such as steering and object tracking~\cite{peng_deeploco_2017}, or improve learning efficiency by reusing low level skills across multiple high level tasks~\cite{frans_meta_2018}. When trained separately, low level controllers can be refined efficiently using prior knowledge, such as by utilising motion capture data~\cite{peng_deepmimic_2018, merel_hierarchical_2019, peng_mcp_2019} or behavioural cloning~\cite{strudel_learning_2020}. For robotics applications it may be difficult to develop controllers for multiple behaviours simultaneously. Using pre-trained primitives can break up a large difficult problem into smaller solvable sub-problems~\cite{schaal_dynamic_2006}.

Using pre-trained policies requires suitable handling of transitions between behaviours. Faloutsos et al.~\cite{faloutsos_composable_2001} learns pre and post-conditions for each controller, such that switching only occurs when these conditions have been satisfied. DeepMimic by Peng et al.~\cite{peng_deepmimic_2018} combines pre-trained behaviours learned from motion capture, with reliance on a phase variable to determine when one behaviour has been completed. Policy sketches introduce a hierarchical method that uses task specific policies, with each task performed in sequence~\cite{andreas_modular_2017}. CompILE uses soft boundaries between task segments~\cite{kipf_compile_2019}. Work by Peng et al.~\cite{peng_terrain-adaptive_2016} trains several actor-critic control policies, modulated by the highest critic value in a given state. For these examples there must be a reasonable state overlap between controllers.

Other methods that combine pre-trained behaviours learn latent representations of skills~\cite{pertsch_accelerating_2020} or primitives~\cite{ha_distilling_2020}, enabling interpolation in the latent space. From an offline dataset of experience, Pertsch et al.~\cite{pertsch_accelerating_2020} were able to combine low level controllers in manipulation tasks and locomotion for a multi-legged agent. Ha et al.~\cite{ha_distilling_2020} utilise motion capture to learn latent representations of primitives, then use model predictive control to navigate with a high dimensional humanoid. The FeUdal approach learns a master policy that modulates
low-level policies using a learned goal signal~\cite{vezhnevets_feudal_2017}. Work by Peng et al.~\cite{peng_mcp_2019} combines pre-trained policies using a gating function that learns a multiplicative combination of actions. In our previous work~\cite{tidd_learning_2021}, we learn when to switch between low level primitives for a simulated biped using data collected by randomly switching between behaviours. Interpolation between behaviours yields natural transitions~\cite{xu_hierarchical_2020}. Yang et al.~\cite{yang_multi_2020} learn a gating neural network to blend several separate expert neural network policies to perform trotting, steering, and fall recovery in real-world experiments with a quadruped. Da et al.~\cite{da_supervised_2017} use supervised learning to train a control policy for a biped from several manually derived controllers to perform periodic and transitional gaits on flat and sloped ground. In each of these approaches, experience from all behaviours must be available during training. 

For dynamic platforms, where separate controllers occupy different subsets of the state, changing behaviours may result in instability if there is no overlap between controllers. Lee et al.~\cite{lee_composing_2019} learn a proximity predictor to train a transition policy to guide an agent to the initial state required by the next controller. Locomotion experiments are demonstrated with a 2D planar walker, where learning occurs by training with all terrain conditions. We show that our method performs more reliably (with a higher success rate), despite training with experience from a single terrain type. Our setup policies learn to transition between complex behaviours with a 3D biped in a simulation environment using a depth image to perceive the terrain. 

\section{Method}
\label{sec:method}

In this section we describe the problem we are solving and our method for training setup policies. 

\begin{figure}[tb!]
\centering
\subfloat[]{\includegraphics[width=\columnwidth]{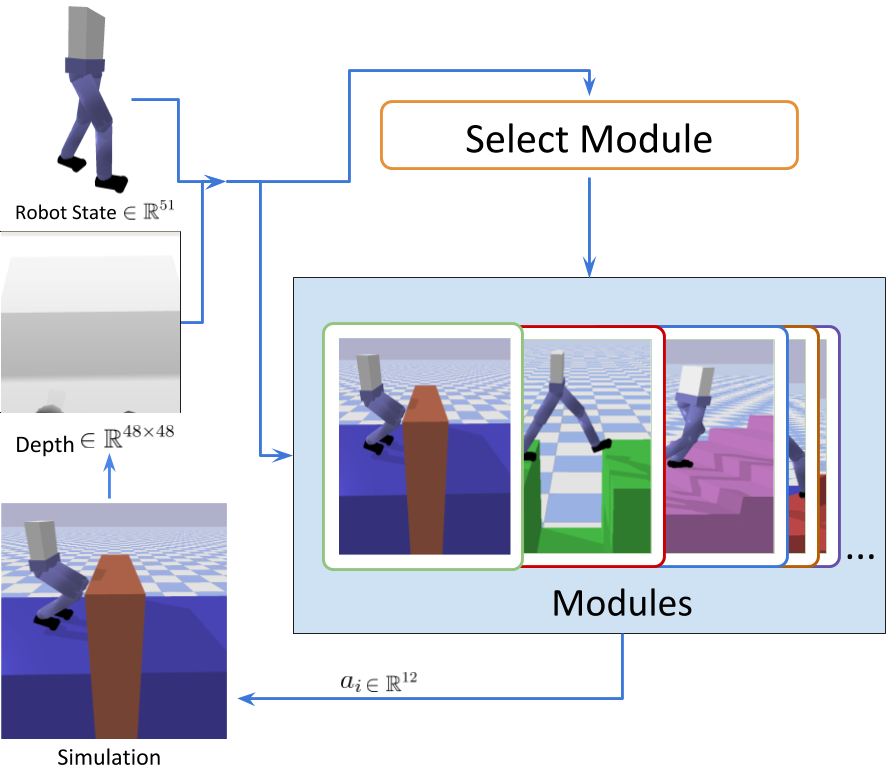}}
\hfill
\subfloat[]{\includegraphics[width=\columnwidth]{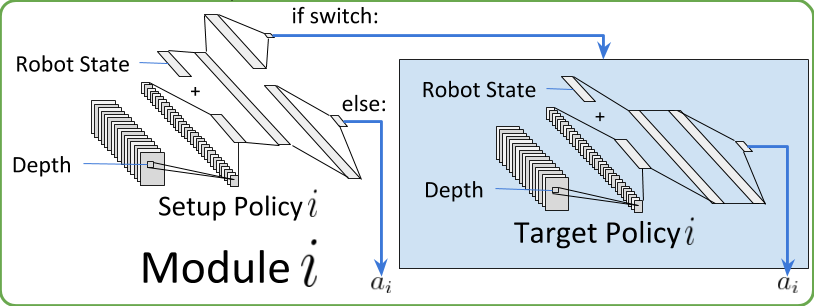}}
\caption{a) shows the modular pipeline. A terrain classifier selects which module should be utilised, and the torque from the output of the selected module is applied to the robot, returning robot state and depth image. b) Each module is pre-trained on a single terrain type. The target policy has been designed to traverse a specific terrain condition. The setup policy guides the trajectory of the robot from a default walking controller to the target policy, also learning when to switch to the target policy.}
\label{fig:fig2}
\end{figure}

\subsection{Problem Description}
\label{sec:problem}

We investigate composing controllers for a dynamic platform with narrow or no state overlap between adjacent behaviours. An example of where this occurs when a bipedal robot performs a jump over a large block. In this scenario, the robot walks up to the block using a walking policy and then performs a jump with a terrain-specific target policy, where the behaviour involves jumping with both feet. Fig~\ref{fig:fig1} shows that the trajectory of the walk policy does not intersect with the trajectory of the target policy, therefore we require an intermediate policy. While it may be possible to improve the robustness of the target policies after they have been trained (i.e. to include states visited by the walk policy), we consider the case where target behaviours and value functions are provided as a black box and are therefore immutable. Our agent has 12 torque controlled actuators, simulated in PyBullet~\cite{coumans_pybullet_2020}.

We select which terrain module to employ using an oracle terrain classifier (Fig~\ref{fig:fig2}a). From a walking policy, we utilise a setup policy that prepares the robot for the target policy (Fig~\ref{fig:fig2}b)). When to switch from the setup policy to the target policy is also a learned output of the setup policy. We first study the traversal of a single difficult terrain artifact that requires walking on a flat surface and performing a jump. We then consider several diverse terrain types (gaps, hurdles, stairs, and stepping stones). The state provided to each policy is $s_t=[rs_t, I_t]$, where $rs_{t}$ is the robot state and $I_t$ is the perception input at time $t$. 

\noindent\textbf{Robot state}: $rs_t=[J_t, Jv_t, c_t, c_{t-1}, v_{CoM,t}, \omega_{CoM,t}, \\ \theta_{CoM,t}, \phi_{CoM,t}, h_{CoM,t}, s_{right}, s_{left} ]$, where $J_t$ are the joint positions in radians, $Jv_t$ are the joint velocities in rad/s, $c_t$ and $c_{t-1}$ are the current and previous contact information of each foot (four Boolean contact points per foot), $v_{CoM,t}$ and $\omega_{CoM,t}$ are the linear and angular velocities of the robot body, $\theta_{CoM,t}$ and $\phi_{CoM,t}$ are the pitch and roll angles of the robot body and $h_{CoM,t}$ is the height of the robot from the walking surface. $s_{right}$ and $s_{left}$ are Boolean indicators of which foot is in the swing phase, and is updated when the current swing foot makes contact with the ground. Robot body rotations are provided as Euler angles. In total there are $51$ elements to the robot state, which is normalised by subtracting the mean and dividing by the standard deviation for each variable (statistics are collected as an aggregate during training).

\noindent\textbf{Perception}: Perception is a depth sensor mounted to the robot base, with a resolution of $[48,48,1]$, and field of view of 60 degrees. Each pixel is a continuous value scaled between $0-1$, measuring a distance between 0.25 and $\SI{2}{\metre}$ from the robot base. The sensor moves with the body in translation and yaw, we provide an artificial gimbal to keep the sensor fixed in roll and pitch. The sensor is pointed at the robot's feet (downwards 60 degrees so the feet are visible) and covers at least two steps in front of the robot~\cite{zaytsev_two_2015}. We reduce the sampling rate of the perception to $\SI{20}{\hertz}$ (reducing computational complexity), where the simulator operates at $\SI{120}{\hertz}$.

\subsection{Deep Reinforcement Learning for Continuous Control}
\label{sec:ppo}
 
We consider our task to be a Markov Decision Process $\text{MDP}$, defined by tuple $\{\mathcal{S},\mathcal{A}, \mathcal{R}, \mathcal{P}, \gamma\}$ where $s_t \in \mathcal{S}$, $a_t \in \mathcal{A}$, $r_t \in \mathcal{R}$ are state, action and reward observed at time $t$, $\mathcal{P}$ is an unknown transition probability from $s_t$ to $s_{t+1}$ taking action $a_t$, and applying discount factor $\gamma$. 

The goal of reinforcement learning is to maximise the sum of future rewards $R = \sum_{t=0}^{T}\gamma^tr_t$, where $r_t$ is provided by the environment at time $t$. Actions are sampled from a deep neural network policy $a_t\sim\pi_\theta(s_t)$, where $a_t$ is the joint commands. Each policy is trained with Proximal Policy Optimisation (PPO)~\cite{schulman_proximal_2017}. We use the implementation of PPO from OpenAI Baselines~\cite{dhariwal_openai_2017}.




\subsection{Training Setup Policies}
\label{sec:setup}

For each target terrain type we train a setup policy with the purpose of bridging the trajectories of a default walking policy and the behaviour for traversing the upcoming terrain. The algorithm for training each setup policy is shown in Algorithm~\ref{alg:setup}. 

\textbf{Default Policy}: 
The default policy is a pre-trained policy designed to walk on a flat surface. When training a setup policy we always begin with the default policy until the terrain artifact is detected by an oracle. Once the terrain has been traversed, we switch back to the default policy when a termination criteria is reached ($\tau_\theta$, defined below). We initialise the setup policy from the default walking policy such that the robot continues to walk towards the obstacle while learning how to prepare for the target policy.

\textbf{Setup Policy}: 
When terrain type $i$ is detected, we switch immediately from the default policy to the setup policy for terrain type $i$, denoted as $\pi_\phi^i$. While training we are given access to an oracle terrain detector that indicates when the robot is a fixed distance from the terrain artifact. The setup policy outputs joint torques $a_i$, and switch condition $\tau_\phi$. If the switch condition is met, we immediately switch to the target policy.

\textbf{Target Policy}: 
The target policy for terrain type $i$, denoted as $\pi_\theta^i$, is trained with curriculum learning following the method outlined in prior work~\cite{tidd_guided_2020}. We also define a termination condition $\tau_\theta$ that denotes switching from the target policy back to the default policy. For the the termination condition to be reached, the robot must successfully traverse the terrain obstacle and have both feet in contact with the ground. This condition is the same for all artifact types.

\textbf{Setup Policy Reward}: 
We design a reward function to motivate the setup policy to transition the robot from states visited by the default policy to states required by the target policy. We note the value function of the target policy, $V^{\pi_\theta^i}(s_t) = \sum_{t=0}^{T}\gamma^tr_t$, (where $r_t$ is the original reward afforded by the environment for the target policy), provides an estimate of what return we can expect if we run the target policy from $s_t$. However, value functions are notoriously over-optimistic for states that have not been visited~\cite{haarnoja_soft_2018} (such as those we might experience by running the default policy).

The advantage is a zero centered calculation of how much better or worse policy $\pi$ performs compared to the value function prediction ${V}^\pi(s_t)$:

\begin{equation}
    A^\pi(s_t, a_t) = R_t - V^\pi(s_t)
\end{equation}

where $R_t$ is sum of future rewards from time $t$ running policy $\pi$.

We estimate the advantage using the temporal difference (TD) error with the value function of the target policy $\pi_\theta^i$:

\begin{equation}
    \hat{A}^{\pi_\theta^i}(s_t, a_t) = r_t + \gamma V^{\pi_\theta^i}(s_{t+1}) - V^{\pi_\theta^i}(s_t)
\end{equation}

where $r_t$ is the target policy reward.

Using advantage $\hat{A}^{\pi_\theta^i}(s_t, a_t)$ as an indication of the accuracy of $V^{\pi_\theta^i}(s_t)$, we define the reward \textbf{Advantage Weighted Target Value}:
\begin{equation}
    \hat{r}_t = ( 1 - \text{min}(\alpha\hat{A}^{\pi_\theta^i}(s_t, a_t)^2,1) ) \cdot \beta\hat{V}^{\pi_\theta^i}(s_t)
    \label{eq:eq1}
\end{equation}

where $\alpha$ and $\beta$ are scaling factors tuned empirically (set to 0.15 and 0.01 respectively). 

The target policy value function $V^{\pi_\theta^i}(s_t)$ has learned the value for states where the target policy has collected experience. Outside of these states, for example when the robot is activating a different policy, we expect the reward provided by the environment $r_t$, and the next state $s_{t+1}$ will not be consistent with what is expected by the target policy value function. Therefore, the advantage $\hat{A}^{\pi_\theta^i}(s_t, a_t)$ will not be close to zero, and $\hat{r}_t$ becomes small. Intuitively, for states where the target policy value function $V^{\pi_\theta^i}(s_t)$ is accurate, the target policy advantage $\hat{A}^{\pi_\theta^i}$ will be close to zero, thus the setup policy will learn to maximise $V^{\pi_\theta^i}$. States where $\hat{A}^{\pi_\theta^i}$ is far from zero will reduce the effect of an overconfident $V^{\pi_\theta^i}$. 

\textbf{Extended Reward}: 
One issue with training setup policies is the delayed effect of actions, where once switched to the target policy, the setup policy is no longer acting on the environment or receiving a reward. Once the setup policy has transitioned to the target policy, we must provide the setup policy with information regarding the performance of the robot since switching. Our method for solving this problem is to provide the additional reward obtained \textit{after} switching, by including an extended reward in the rollout buffer of the setup policy. We add this extended reward to the last reward received running the setup policy: 
\begin{equation}
    \hat{r}_{final} = \hat{r}_{final} + \hat{r}_t 
\end{equation}
where $\hat{r}_{final}$ is the final reward entry received by the setup policy before switching, and $\hat{r_t}$ is the reward received at time $t$, after the setup policy has transitioned to the target policy. For algorithmic stability, we clear all but the last entry of the buffer after each training update. The last reward entry of the buffer becomes $\hat{r}_{final}$ for environment steps that continue after training. The procedure for training setup policies is provided in Algorithm~\ref{alg:setup}. Note that the termination variable $\tau_t$ (line 7) is either $\tau_\phi$ if the current policy is the setup policy, or $\tau_\theta$ if the current policy is the target policy.

\begin{algorithm}
	\caption{Setup Before Switching} 
	\begin{algorithmic}[1]
      \State Load Target Policy $\pi_\theta^i$. Load Default Policy $\pi_\theta^j$
      \State Initialise Setup Policy $\pi_\phi^i$ from the Default Policy
      \State Initialise Buffer
        \State Current Policy $\gets$ Default Policy
		\For {$iteration=1,2,\ldots$}
		    \While {not $done$}
    		    \State $a_t, \tau_t \sim $ Current Policy$(s_t)$ \Comment{$\tau$ is $\tau_\phi$ or $\tau_\theta$}
    		    \State $s_{t+1}, r_t, done = env.step(a_t)$
    		    \If{Terrain detected}
    	            \State $\hat{A}^{\pi_\theta^i}(s_t, a_t) = r_t + \gamma V^{\pi_\theta^i}(s_{t+1}) - V^{\pi_\theta^i}(s_t)$
        		    \State $\hat{r}_t = (1 - \text{min}(\alpha\hat{A}^{\pi_\theta^i}(s_t, a_t)^2, 1))\cdot \beta V^{\pi_\theta^i}(s_t)$
        		    \If{Current Policy == Default Policy}
        		        \State Current Policy $\gets$ Setup Policy $\pi_\phi^i$
        		        \State $\tau_t \gets \tau_\phi$
            		\ElsIf{Current Policy == Setup Policy}
        		        \State Store $s_t$, $a_t$, $\hat{r}_t$, $\tau_\phi$ into Buffer
                        \If{Buffer Full}
                            \State PPO Training Update
                            \State Clear buffer (except last entry)
                        \EndIf
            		    \If {$\tau_\phi$}
            		        \State Current Policy $\gets$ Target Policy $\pi_\theta^i$
            		        \State $\tau_t \gets \tau_\theta$
            		    \EndIf
        			\ElsIf{$\tau_\theta$}
        			    \State Current Policy $\gets$ Default Policy $\pi_\theta^j$
        		    \EndIf
        		    \If{Current Policy != Setup Policy}
        			    \State $\hat{r}_{final} = \hat{r}_{final} + \hat{r}_t$ 
        			\EndIf
        		\EndIf
        		\State $s_t = s_{t+1}$
        	\EndWhile
		\EndFor
	\end{algorithmic} 
\label{alg:setup}
\end{algorithm}


\section{Experiments}
\label{sec:experiments}

We evaluate our method with a biped in simulation. Initial evaluations are performed with a single difficult terrain (Sec~\ref{sec:ablation}, Sec~\ref{sec:reward_choice}, Sec~\ref{sec:comparisons}), before adding more behaviours in Sec~\ref{sec:multi_terrain}. The difficult terrain is a sample of flat terrain with a single block 50 cm high, and 30 cm in length (in the forward direction of the robot), an example is shown in Fig.~\ref{fig:fig1}. We perform training and evaluation on a single terrain sample, starting the robot from a random $x$ (forward direction) uniformly sampled from (0.0, 2.2) $\si{\meter}$ before the terrain artifact, $y$ (strafe) uniformly sampled from (-0.6, 0.6) $\si{\meter}$, and $heading$ (yaw) uniformly sampled from (-0.3, 0.3) $\si{\radian}$. All experiments using a single terrain type are evaluated by pausing training and performing 100 evaluation runs, and recording both the Success percentage and Distance percentage. Success \% is defined as the average success rate to reach a goal location several meters past the terrain artifact. Distance \% is the average percentage of total terrain distance the agent successfully covered.

\subsection{Effect of Using Initialisation and Extended Reward}
\label{sec:ablation}

We perform an ablation to study the effect of initialising the setup policy from the walking policy, and the effect of extending the reward after the setup policy has transitioned to the target policy. In all experiments the default policy is used until the terrain is detected, then the setup policy is activated until the termination signal $\tau_\phi$ indicates switching to the target policy, and $\tau_\theta$ then results in switching back to the default policy.

\begin{itemize}[leftmargin=*]

\item\textbf{Without Initialisation:} We evaluate our method without initialising the setup policy from the default walking policy, i.e. the setup policy is randomly initialised.

\item\textbf{Without Extended Reward:} We evaluate our method without including the reward after the setup policy has switched to the target policy. 

\item\textbf{Full Method:} Our method uses a setup policy initialised from the default walk policy, and receives reward signal after the setup policy has transitioned to the target policy. 

\end{itemize}

\begin{table}[h!]
    \vspace{-2mm}
    \centering
    \begin{adjustbox}{min width=0.8\columnwidth}
    \begin{tabular}{cccccc}
                                      & Success \% & Distance \% \\
       \hline
        Without Initialisation        & 38.8      & 73.2  \\
        Without Extended Reward       & 12.3      & 58.3  \\
        Full Method                          & \textbf{82.2} & \textbf{96.1}\\
    \hline
    \end{tabular}
    \end{adjustbox}
    \caption{Ablation study for initialising setup policies from a default walk policy, and receiving an extended reward, for switching from a walking policy to a jump policy on a single terrain sample.}
    \label{tab:results}
\end{table}

We can see that initialising the setup policy with the default walking policy and extending the reward improves learning outcomes as shown in Table~\ref{tab:results}, where our method achieves 82.2\% success compared to 38.8\% without initialisation from the default walking policy and 12.3\% without receiving the extended reward. 

\subsection{Setup Policy Reward Choice}
\label{sec:reward_choice}

\begin{figure}[tb!]
\centering
\includegraphics[width=\columnwidth]{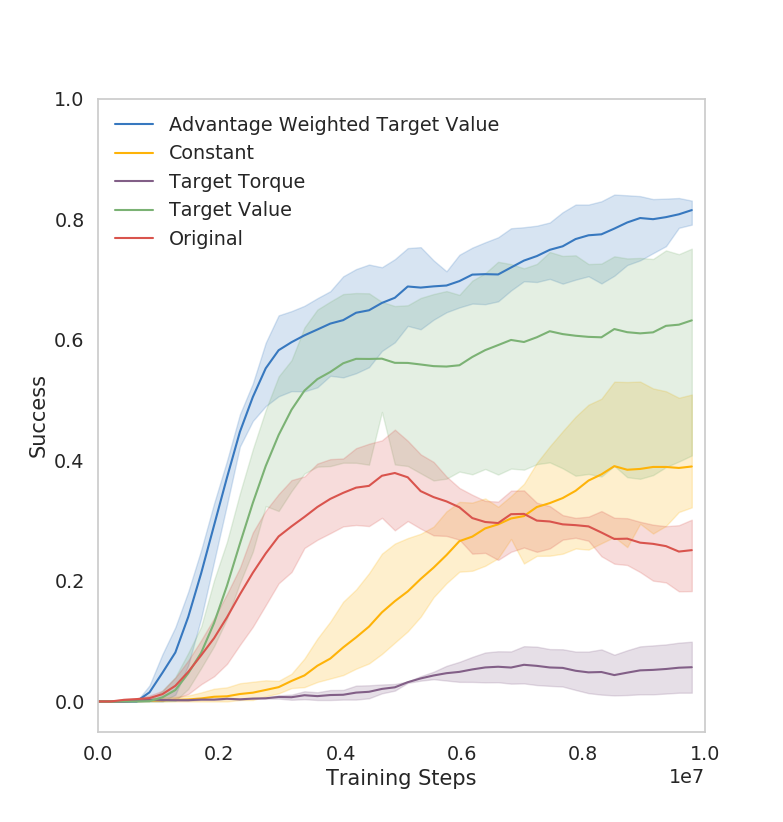}
\hfill
\caption{We investigate several options for the reward function used to train a setup policy on the difficult jump terrain. We show the success rate (robot reaches several meters after the jump) during training for three different random seeds.}
\label{fig:results}
\end{figure}

For training the setup policy, we investigate several options for the reward provided at each timestep. For each reward type we evaluate with three different random seeds, results are shown in Fig~\ref{fig:results}.

\begin{itemize}[leftmargin=*]
\item\textbf{Original:} The reward afforded by the environment that is used to train the target policy: the reward encourages following an expert motion. 

\item\textbf{Constant:} At each timestep a constant reward of 1.5 is given.

\item\textbf{Target Torque:} We encourage the setup policy to match the output of the target policy by minimising the error: $\exp[-2.0\cdot (\pi_\phi(s_t) - \pi_\theta(s_t))^2]$. 

\item\textbf{Target Value:} We evaluate the effect of using the target policy value function as the reward for the setup policy. This is not the same as the original reward, as it is not provided by the environment, but by $\beta V^{\pi_\theta^i}(s_t)$, where $\beta$ is scaling factor 0.01.

\item\textbf{Advantage Weighted Target Value:} We evaluate the reward introduced in Equation \ref{eq:eq1}.

\end{itemize}

It can be seen from Fig~\ref{fig:results} that a reward using \textbf{Advantage Weighted Target Value} performs better than other reward choices, achieving the highest average success rate (78.7\%). Using the target value reaches a success rate of 63.9\%. This disparity, and the large variation in the target value (green line), shows the effect of over estimation bias often seen with value functions, validating our idea that weighting the value by the advantage reduces the effect of this bias when training setup policies.

\subsection{Comparisons}
\label{sec:comparisons}
We compare our method with several others, including end to end methods, without using setup policies, and a method that uses proximity prediction to train a transition policy~\cite{lee_composing_2019}.

\begin{table}[h!]
    \vspace{-2mm}
    \centering
    \begin{adjustbox}{min width=0.8\columnwidth}
    \begin{tabular}{cccccc}
                                      & Success\% & Distance\% \\
       \hline
        Single Policy                 & 0.0           & 11.3  \\
        Single Policy With Curriculum & 0.0           &  3.8 \\
        No Setup Policy               & 1.5           & 51.3 \\
        Learn When to Switch          & 0.0           & 44.2  \\
        Proximity Prediction         & 51.3           & 82.4 \\
        Setup Policy (Ours)         & \textbf{82.2} & \textbf{96.1}\\
    \hline
    \end{tabular}
    \end{adjustbox}
    \caption{Success and average distance of the total length covered when switching from a walking policy to a jump policy on a single terrain sample.}
    \label{tab:results_compare}
\end{table}

\begin{itemize}[leftmargin=*]

\begin{figure*}[tb!]
\centering
\includegraphics[width=\textwidth]{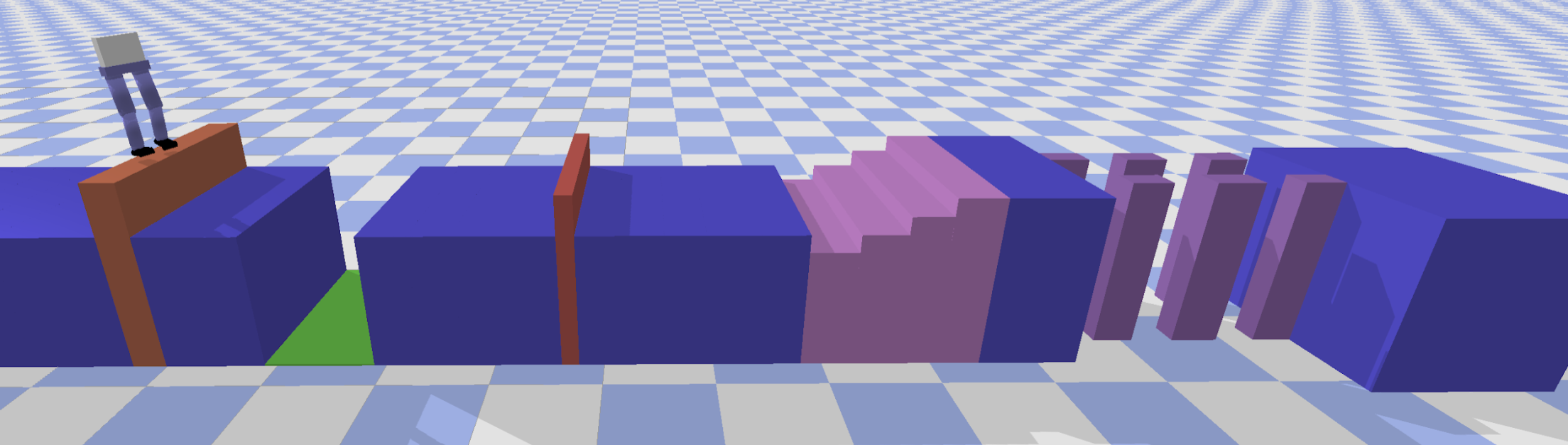}
\hfill
\caption{Setup policies enable a biped to transition between complex visuo-motor behaviours for traversing a sequence of diverse terrains.}
\label{fig:multi}
\end{figure*}

\item\textbf{Single Policy:} We train a single end to end policy on the terrain sample.

\item\textbf{Single Policy With Curriculum:} We utilise curriculum learning to train a single end to end policy on the terrain sample (using the method outlined in~\cite{tidd_guided_2020}).

\item\textbf{No Setup Policy:} We switch to the target policy from the default policy as soon as the terrain oracle detects the next terrain.

\item\textbf{Learn When to Switch:} We collect data by switching from the default policy to the target policy at random distances from the Jump. We use supervised learning to learn when the robot is in a suitable state to switch. Details of this method is found in our previous work~\cite{tidd_learning_2021}.

\item\textbf{Proximity Prediction:} We follow the method defined by Lee et al.~\cite{lee_composing_2019} to train Transition Policies (one for each behaviour) using a proximity predictor function $P(s_t)$. $P(s_t)$ outputs a continuous value that indicates how close the agent is to a configuration that resulted in successful traversal. $P(s_t)$ is trained using supervised learning from success and failure buffers. The reward used to train the transition policy is the dense reward created by  $P(s_{t+1}) - P(s_{t})$, encouraging the agent to move closer to a configuration that results in successful switching. For accurate comparison we initialise the transition policy with weights from the walk policy, and utilise the terrain oracle for policy selection (the paper refers to a rule-based meta-policy in place of a terrain oracle~\cite{lee_composing_2019}).

\item\textbf{Setup Policy (Ours):} We follow Algorithm~\ref{alg:setup} to train setup policies. 

\end{itemize}

We can see from Table~\ref{tab:results_compare} that our method for developing setup policies performs the best (82.2\% success rate), compared to other methods. A single policy was unable to traverse the difficult jump terrain, even with an extended learning time and curriculum learning. The poor performance of \textbf{Single Policy With Curriculum} was the result of the robot not progressing through the curriculum, and as a consequence is unable to move without assistive forces. We show that setup policies are necessary for this problem, with \textbf{No Setup Policy} unable to successfully traverse the terrain (1.5\% success). \textbf{Learning When to Switch} also performed poorly as there are very few states that overlap between the default policy and the target policy for the switch estimator to learn. The \textbf{Proximity Prediction} method was able to successfully traverse the jump 51.3\% of the time. It is worth noting that this method required approximately three times more training episodes than our method to reach the provided results, despite also being initialised from the default policy.

These experiments show that for the difficult jump terrain, we require a specialised target policy dedicated to learning the complex jumping behaviour. We show that this specialised policy has a trajectory that does not overlap with the default policy, and thus setup policies are required. Our method for learning a setup policies achieved the highest performance on this difficult task.

\subsection{Multiple Terrain Types}
\label{sec:multi_terrain}

For our final experiment we train setup policies for several additional terrain types. In total we now have modules for jumps, gaps, hurdles, stairs, and stepping stones. Fig~\ref{fig:multi} shows a sequence with each terrain type. We follow the pipeline shown in Fig~\ref{fig:fig2}a), where a terrain oracle determines which module should be selected. On selection, the setup policy is employed until the robot reaches a switch state ($\tau_\phi$), then switching to the target policy.
Termination criteria $\tau_\theta$ (introduced in Sec~\ref{sec:method}) determines when to switch back to the default policy for the jump obstacle only. All other policies resemble the behaviour of the default policy on flat terrain, i.e. we continue using the target policy after the terrain has been traversed for gaps, hurdles, stairs, and stepping stones, until the oracle determines the terrain has changed.

We evaluate with a sequence of each of the 5 terrain types, randomly shuffled before each run. We perform 1000 runs, with a total of 5000 various terrain types to be traversed. In Table~\ref{tab:multi_results} we show the average percentage of total distance travelled and success rate with and without using setup policies. A successful run in this scenario is defined as traversing the last terrain in the sequence. Table~\ref{tab:multi_breakdown} shows the failure count for each terrain type.

\begin{table}[h!]
    \vspace{-2mm}
    \centering
    \begin{adjustbox}{min width=0.8\columnwidth}
    \begin{tabular}{cccccc}
                                      & Success \% & Distance \% \\
       \hline
        Without Setup Policies        & 1.9      & 36.3  \\
        With Setup Policies       & \textbf{71.2}      & \textbf{80.2}  \\
    \hline
    \end{tabular}
    \end{adjustbox}
    \caption{Success rate and percentage of the total terrain length travelled from 1000 episodes of all 5 terrain types, randomly shuffled each episode.}
    \label{tab:multi_results}
\end{table}

\begin{table}[h!]
    \vspace{-2mm}
    \centering
    \begin{adjustbox}{min width=0.9\columnwidth}
    \begin{tabular}{cccccc}
                                      & Jump & Gap & Hurdle & Stairs & Steps \\
      \hline
        Without Setup Policies       & 782      & 52  & \textbf{36} &  \textbf{31} & \textbf{86}  \\
        With Setup Policies        & \textbf{36}      & \textbf{48}  & 43 & 65 & 96 \\
    \hline
    \end{tabular}
    \end{adjustbox}
    \caption{Number of failures by terrain type from 1000 episodes of all 5 terrain types, randomly shuffled each episode.}
    \label{tab:multi_breakdown}
\end{table}

\begin{table}[h!]
    \vspace{-2mm}
    \centering
    \begin{adjustbox}{min width=0.8\columnwidth}
    \begin{tabular}{cccccc}
                                      & Success \% & Distance \% \\
       \hline
        Without Setup Policies        & 68.2      & 78.1  \\
        With Setup Policies       & \textbf{76.7}      & \textbf{84.6}  \\
    \hline
    \end{tabular}
    \end{adjustbox}
    \caption{Success rate and percentage of the total terrain length travelled from 1000 episodes of 4 terrain types \textbf{(without the jump terrain)}, randomly shuffled each episode}
    \label{tab:multi_results_no_jumps}
\end{table}

\begin{table}[h!]
    \vspace{-2mm}
    \centering
    \begin{adjustbox}{min width=0.82\columnwidth}
    \begin{tabular}{ccccc}
                                      & Gap & Hurdle & Stairs & Steps \\
      \hline
        Without Setup Policies       & 74  & \textbf{41} &  64 & 142  \\
        With Setup Policies        &  \textbf{44}  & 51 & \textbf{49} & \textbf{90} \\
    \hline
    \end{tabular}
    \end{adjustbox}
    \caption{Number of failures by terrain type from 1000 episodes of 4 terrain types \textbf{(without the jump terrain)}, randomly shuffled each episode.}
    \label{tab:multi_breakdown_no_jumps}
\end{table}
\newpage
We can see from Table~\ref{tab:multi_results} that setup policies improve the success rate compared to switching policies without the intermediate policy (71.2\% success compared to 1.9\%). Table~\ref{tab:multi_breakdown} provides a breakdown of the number of times the robot failed on each terrain. From Table~\ref{tab:multi_breakdown} we can see that setup policies are most critical for traversing the difficult jump terrain (reducing the failures from 782 to 36), though we also see improvements for gap terrain. 

To investigate further, we exclude the jump obstacle from the terrain sequence and still observe a significant benefit using setup policies. Table~\ref{tab:multi_results_no_jumps} shows the performance increase from 68.2\% to 76.7\% of successful traversals for 4 terrain types (excluding the jump obstacle). The stepping stone obstacle had the highest failure rate of the remaining terrain types, investigation is required to improve the setup policies for this terrain (Table~\ref{tab:multi_breakdown_no_jumps}).

Despite the clear need for setup policies, we attribute failures to out of distribution states, i.e. the robot visits states in the terrain sequence that it does not experience when training the setup policy on the individual terrain types. Training setup policies for wider range of states will be investigated in future work. Furthermore, the performance of behaviours on each individual terrain type could be analysed to improve the generalisation of this method to new behaviours.




\section{Conclusion}
\label{sec:conclusion}

It is common practice for legged robots to have separate locomotion policies to traverse different terrain conditions. We propose setup policies that enable smooth transition from the trajectory of one locomotion policy to the trajectory of a target policy for traversing difficult terrain with a dynamic biped. We use deep reinforcement learning to learn the setup policies for which we introduce a novel reward based on the Advantage Weighted Target Value, utilising the scaled value prediction of the target policy as the reward for the setup policy. 

In simulation experiments for transitioning from a walking policy to a jumping policy, we show using an in-between setup policy yields a higher success rate compared to using a single policy (0$\%$ success rate), or the two policies without the setup policy (from 1.5$\%$ success rate without setup policies to 82$\%$). We further show that our method is scalable to other terrain types, and demonstrate the effectiveness of the approach on a sequence of difficult terrain conditions, improving the success rate from 1.9$\%$ without setup policies to 71.2$\%$.

A limitation of our method is that setup policies are trained from a default walking policy. We would like to improve the scope of the setup policies such that a wider range of states are funnelled towards the target policy. If we allowed further training of the target policies, our training method could be used to include more states and improve robustness. The idea of utilising the advantage (estimated by the temporal difference (TD) error) as a confidence metric can be applied generally for region of attraction (RoA) expansion, safe reinforcement learning, and also for blending several behaviours for performing complex composite tasks. These ideas will be explored in future work. A significant assumption in this work is that we have access to a terrain oracle, removing this dependency will required as we consider experiments in the real world.

\addtolength{\textheight}{-2cm}   
\bibliographystyle{IEEEtran}
\bibliography{references}
\end{document}